\documentclass[letterpaper, 10 pt, conference]{ieeeconf}

\usepackage{amsmath,amsfonts}
\usepackage{algorithm}
\usepackage{array}
\usepackage{textcomp}
\usepackage{stfloats}
\usepackage{url}
\usepackage{verbatim}
\usepackage{graphicx}
\usepackage{cite}
\usepackage{xcolor}
\usepackage{adjustbox}
\usepackage[symbol]{footmisc}
\usepackage{makecell}
\usepackage{times}

\usepackage{multicol}

\usepackage{amsmath}
\usepackage{amssymb}
\usepackage{cases}
\usepackage{graphicx}
\usepackage{booktabs}

\usepackage{amsthm}

\theoremstyle{definition}

\usepackage[]{xcolor}
\usepackage{subcaption}
\usepackage{wrapfig}
\usepackage{multirow}
\usepackage{mathtools}
\usepackage{empheq}
\usepackage{color}
\usepackage{bbm}
\usepackage{xspace}
\let\labelindent\relax
\usepackage[shortlabels]{enumitem}

\usepackage{algorithmicx}
\usepackage{algorithm}
\usepackage[]{algpseudocode}

\usepackage{fontawesome}

\usepackage{soul}
\usepackage{comment}

\usepackage[]{url}

\usepackage[pdftex, colorlinks=true, linkcolor=black, citecolor = black, urlcolor = blue, filecolor=black , pagebackref=false, hypertexnames=false, bookmarks=false]{hyperref}
\usepackage{cleveref}
\crefname{problem}{Problem}{Problems}
\crefname{example}{Example}{Examples}
\crefname{section}{Sec.}{Secs.}
\Crefname{section}{Section}{Sections}
\Crefname{table}{Table}{Tables}
\crefname{table}{Table}{Tabs.}
\crefname{figure}{Fig.}{Figs.}
\crefname{algorithm}{Algorithm}{Algorithms}
\crefname{remark}{Remark}{Remarks}
\crefname{theorem}{Theorem}{Theorems}
\crefname{lemma}{Lemma}{Lemmas}
\crefname{corollary}{Corollary}{Corollaries}
\crefname{assumption}{Assumption}{Assumptions}
\usepackage[font=small]{caption}

\usepackage{siunitx}

\newcommand{\bmat}{\begin{bmatrix}}
\newcommand{\emat}{\end{bmatrix}}

\newcommand{\eg}{\emph{e.g.}\xspace}

\providecommand{\optional}[1]{{}}

\providecommand{\techreport}[1]{{}}

\newcommand{\M}[1]{{\mathbf #1}} %
\newcommand{\ve}[1]{{\mathbf #1}} %

\newcommand{\R}{{\mathbb R}}

\makeatletter
\newcommand\notsotiny{\@setfontsize\notsotiny\@vipt\@viipt}
\makeatother

\definecolor{neonblue}{rgb}{0.1, 0.7, 1}
\definecolor{neongreen}{rgb}{0.1, 1, 0.1}

\definecolor{todocolor}{HTML}{7F3DB8}
\definecolor{citecolor}{HTML}{F64A8A}

\newcommand{\GM}{\text{GM}}

\newcommand{\objaerial}{o^\text{a}}
\newcommand{\objground}{o^\text{g}}

\newcommand{\lc}{\mathbf{T}^{m,n}_{k,\ell}}
\newcommand{\lcdist}{\mathcal{T}^{m,n}_k}

\newcommand{\Tground}[1]{\M{X}_{#1}}
\newcommand{\Taerial}[1]{\M{A}_{#1}}

 \newcommand{\FGsq}{FG$^2$} 

\definecolor{tabblue}{HTML}{1F77B4}
\definecolor{taborange}{HTML}{FF7F0E}
\definecolor{tabgreen}{HTML}{2CA02C}
\definecolor{tabred}{HTML}{D62728}
\definecolor{tabpurple}{HTML}{9467BD}
\definecolor{tabbrown}{HTML}{8C564B}
\definecolor{tabpink}{HTML}{E377C2}
\definecolor{tabgray}{HTML}{7F7F7F}
\definecolor{tabolive}{HTML}{BCBD22}
\definecolor{tabcyan}{HTML}{17BECF}

\IEEEoverridecommandlockouts
\begin{document}

\title{\LARGE Meridian: Metric-Semantic Primitive Matching for Cross-View Geo-Localization Beyond
Urban Environments}

\author{
\begin{tabular}{c}
Mason Peterson\textsuperscript{*,1},
Qingyuan Li\textsuperscript{*,1},
Yixuan Jia\textsuperscript{1},
Fernando Cladera\textsuperscript{2}, \\
Carlos Nieto-Granda\textsuperscript{3},
Camillo Jose Taylor\textsuperscript{2},
Jonathan P. How\textsuperscript{1}
\end{tabular}
\thanks{This work is supported by ARL DCIST under Cooperative Agreement Number W911NF-17-2-0181 and DSTA.}
\thanks{\textsuperscript{*}These authors contributed equally to this work.}
\thanks{\textsuperscript{1}Massachusetts Institute of Technology, Cambridge, MA 02139, USA. \{\texttt{masonbp, andyli27, yixuany, jhow}\}\texttt{@mit.edu}.}
\thanks{\textsuperscript{2}GRASP Laboratory, University of Pennsylvania, Philadelphia, PA 19104, USA. \{\texttt{fclad, cjtaylor}\}\texttt{@seas.upenn.edu}.}
\thanks{\textsuperscript{3}U.S. Army Combat Capabilities Development Command, Army Research Laboratory, Adelphi, MD 20783, USA. \texttt{carlos.p.nieto2.civ@army.mil}.}
}

\maketitle

\begin{abstract}
Successful robot automation requires accurate global localization to support repeatability, task planning, goal specification, and safe operation. 
However, reliable localization in GNSS-denied environments remains an open problem. 
Overhead aerial imagery offers a promising solution, but existing approaches primarily target structured urban environments and have been rarely demonstrated in unstructured natural terrain.
Limitations of the state-of-the-art include a reliance on models trained for specific environments, as well as difficulty handling repetitive geometries and featureless landscapes commonly found in natural outdoor areas.
To overcome these challenges, we present Meridian, a method for matching high-level metric-semantic primitives across aerial images and ground robot RGB-D camera data that achieves accurate global localization and generalizes well across diverse environments, all without any training or algorithmic fine-tuning on area-specific data.
We formulate novel consistency metrics to estimate a distribution over robot submap poses and to reject outlier hypotheses in a robust pose graph optimization step for accurate robot trajectory estimation.
We demonstrate that our algorithm can localize a ground robot across a wide variety of environments, including an autonomous driving dataset, a park and campus area, and a wilderness camp, with an average optimized trajectory error of 2.4 m over 19 km of ground traversal.
\end{abstract}

\section{Introduction}
\label{sec:intro}

Consistent robot localization in a global coordinate frame is a foundational prerequisite for downstream autonomy tasks~\cite{sh-p1-prelude}. 
Aerial imagery acquired from satellites~\cite{kim2017satellite,wang2023satellite} or drones~\cite{miller2021any} has shown significant utility for consistent pose estimation.
Large-scale aerial images can serve as a globally consistent reference for a ground robot, allowing the robot to match features in its local sensor frame with features in the aerial image, enabling \emph{cross-view geo-localization}---estimation of the ground robot's pose with respect to the global coordinate frame of the aerial image.

Localizing in an aerial image offers many benefits in downstream autonomy tasks.
For example, knowing a robot's pose in an aerial image gives a human operator easy situational awareness, serving as a means for the operator to specify mission plans and for the robot to communicate task performance results~\cite{hsieh2007adaptive}. 
Additionally, cross-view localization can be useful for multi-robot collaboration, enabling robots to share maps and plans without requiring robots to ever be near each other.
The utility offered by cross-view localization has made it a well-studied problem; however, localization in unstructured natural environments remains comparatively underexplored. Most state-of-the-art methods instead focus on autonomous driving applications in urban settings, where roads, buildings, and other structured features provide strong geometric cues for localization \cite{xia2025fg,zhang2024increasing,wang2024view,wang2023satellite}.
Localizing from aerial imagery in forested or mountainous environments introduces significant challenges, including repetitive geometries and featureless landscapes, which make localization difficult. 
Furthermore, most current cross-view localization works rely on neural networks that have been trained in areas similar to the region of operation~\cite{shi2020beyond,xia2025fg}, and suffer from decreased performance in novel regions.

\begin{figure}[t!]
    \centering
    \includegraphics[width=\columnwidth]{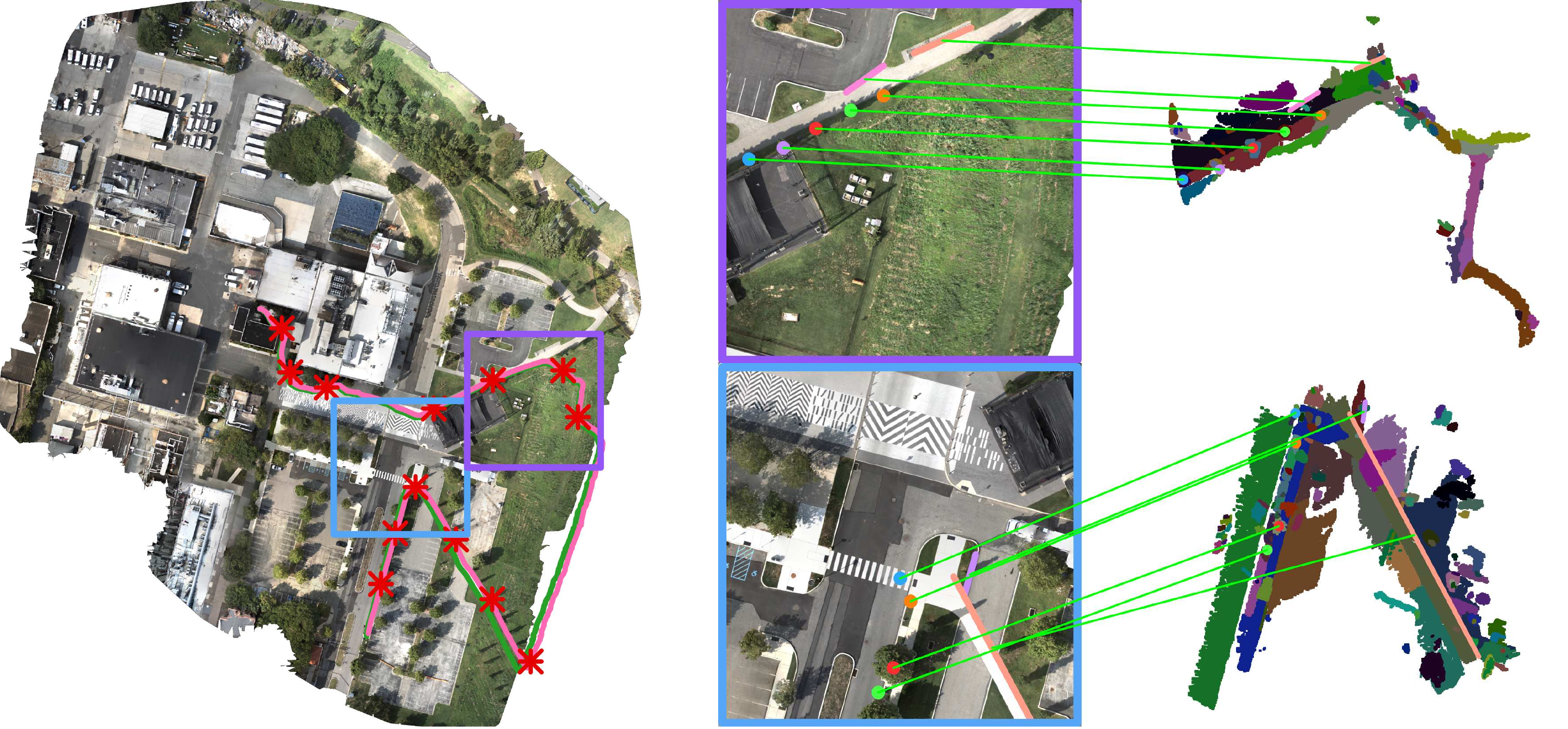}
    \caption{Our algorithm maps and matches object centroids and region borders across aerial and ground views, intuitively representing abstract elements of an environment that are key indicators for localization.
    We propose a robust data association method that leverages semantic and geometric cues to find correct associations, enabling cross-view geo-localization in diverse environments.
    }
    \label{fig:intro-fig}
    \vspace{-15pt}
\end{figure}

In contrast, our work presents a method for performing cross-view geo-localization in general environments without requiring any prior knowledge of the environment, enabling easy generalization.
We leverage foundation models~\cite{zhao2023fast,kirillov2023segment,oquab2023dinov2} for front-end open-set extraction of points (representing objects) and line segments (representing area borders) in any environment.
We propose a novel mechanism for extracting hypotheses of point and line matches and incorporate these hypotheses into an outlier rejection framework for accurate cross-view geo-localization.
We call our proposed method \emph{Meridian} for its metric-semantic primitive data association.
We summarize our contributions as follows:
\begin{enumerate}
    \item A method for extracting point and line primitives from aerial and ground views as inputs to a sparse matching problem for cross-view geo-localization.
    \item A data association distance metric that enables correct primitive matching in extreme outlier regimes.
    \item A robust pose graph optimization algorithm that incorporates submap registration measurements as probability distributions and solves for a set of inlier measurements via discrete outlier measurement rejection.
    \item A unified cross-view localization system operating on RGB-D inputs that demonstrates the ability to localize a robot in a variety of challenging environments and across different seasons without any fine-tuning or pose initialization.
\end{enumerate}

The code and Camp Dataset associated with the paper will be made open-source.

\section{Related Work}
\label{sec:literature}

Research in cross-view geo-localization spans both coarse, region-level position estimation (on the order of 10--100\,m) and meter or sub-meter-level fine-grained pose estimation.

\noindent 
\textbf{Coarse Cross-View Image Matching.}
Early cross-view localization work formulated air-ground localization as a database retrieval problem, where the goal was to identify the best matching aerial image for a given ground image query~\cite{lin2013cross,durgam2024cross}. Initial approaches relied on hand-crafted descriptors to establish correspondences between aerial and ground imagery. More recently, Siamese-network-based methods have demonstrated strong retrieval performance by learning global descriptors for aerial and ground images within a shared embedding space~\cite{hu2018cvm,downes2022city}.

\noindent 
\textbf{Fine-Grained Cross-View Pose Estimation.}
Recent advances in learned architectures and training strategies have enabled neural networks to achieve fine-grained, meter-level pose estimation from ground imagery~\cite{shi2020beyond}, although many such methods assume known orientation. 
More recent work explores hybrid learning and feature matching approaches~\cite{xia2025fg,xia2025revisiting}, as well as the incorporation of domain priors, such as exploiting the greater co-visibility of structures with higher elevation~\cite{wang2024view}. 
Fervers et al.~\cite{fervers2023uncertainty} 
evaluate a dense grid of candidate poses to estimate a probability distribution over poses rather than a single hypothesis.

\noindent 
\textbf{Cross-View Robot Trajectory Estimation.}
Estimating a robot’s full trajectory in an aerial image extends fine-grained cross-view pose estimation to trajectory estimation over time.
These works can be categorized into algorithms that assume a vehicle begins with a coarse initial pose estimate and focus on minimizing drift and estimation error using overhead imagery~\cite{zhang2024increasing,munoz2024geo}, and those that tackle both global localization in an aerial image and fine-grained pose estimation after convergence~\cite{miller2021any,ankenbauer2023global}.
Approaches to cross-view trajectory estimation include incorporating likelihood measurements of potential poses into a particle filter~\cite{kim2017satellite,miller2021any} and incorporating fine-grained pose measurements into a factor graph~\cite{zhang2024increasing}.
Miller et al.~\cite{miller2021any} propose an algorithm for extracting coarse semantics from a LiDAR point cloud and efficiently scoring potential poses based on polar class-wise images.
Most similar to our work, Ankenbauer et al.~\cite{ankenbauer2023global} detect and match predefined objects (rocks and cars) across views.
Adjacent to cross-view localization, \cite{frosi2023osm,przewodowski2024global,li2026odometry,lee2024autonomous} use OpenStreetMap (OSM) for building and road constraints to guide accurate localization.
In comparison, we use only a single limited-field-of-view camera with depth values (which can also be obtained through LiDAR-RGB fusion) and an aerial orthophoto, without requiring scene-specific training 
or a prior map.

\noindent 
\textbf{Geometric Feature Matching.}
Similar to Ankenbauer et al.~\cite{ankenbauer2023global}, we leverage the geometric similarity of observed features across aerial and ground views for matching and subsequent pose estimation.
Geometry-based correspondence finding has roots in global point cloud registration, which seeks to register two point clouds by finding associations between pairs of points~\cite{zhou2016fast}.
Methods for performing global point cloud registration include pose estimation via graduated non-convexity optimization~\cite{zhou2016fast,yang2020teaser}, random consensus sampling~\cite{ransac}, and optimizing for large cliques of mutually consistent associations~\cite{lusk2024clipper}.
Maximum clique or densest subgraph formulations have been used for performing global registration of 3D objects by representing objects by their centroids and leveraging geometric consistency for matching across views~\cite{dube2018incremental,ankenbauer2023global,peterson2025roman}.
Recent work has formulated line and plane primitive matching using the Grassmannian metric to measure distances across pairs of primitives for consistency scoring~\cite{lusk2022graffmatch,yu2025slim}.
However, the Grassmannian metric can only represent primitives as unbounded, meaning line segments cannot be differentiated from infinite lines.
In this work, we propose a novel scoring method that accounts for bounded lines and can be used with a known rotation or translation to guide the solution when available.

\section{Proposed Approach}
\label{sec:method}

\begin{figure}[t]
    \centering
    \includegraphics[width=0.42\textwidth]{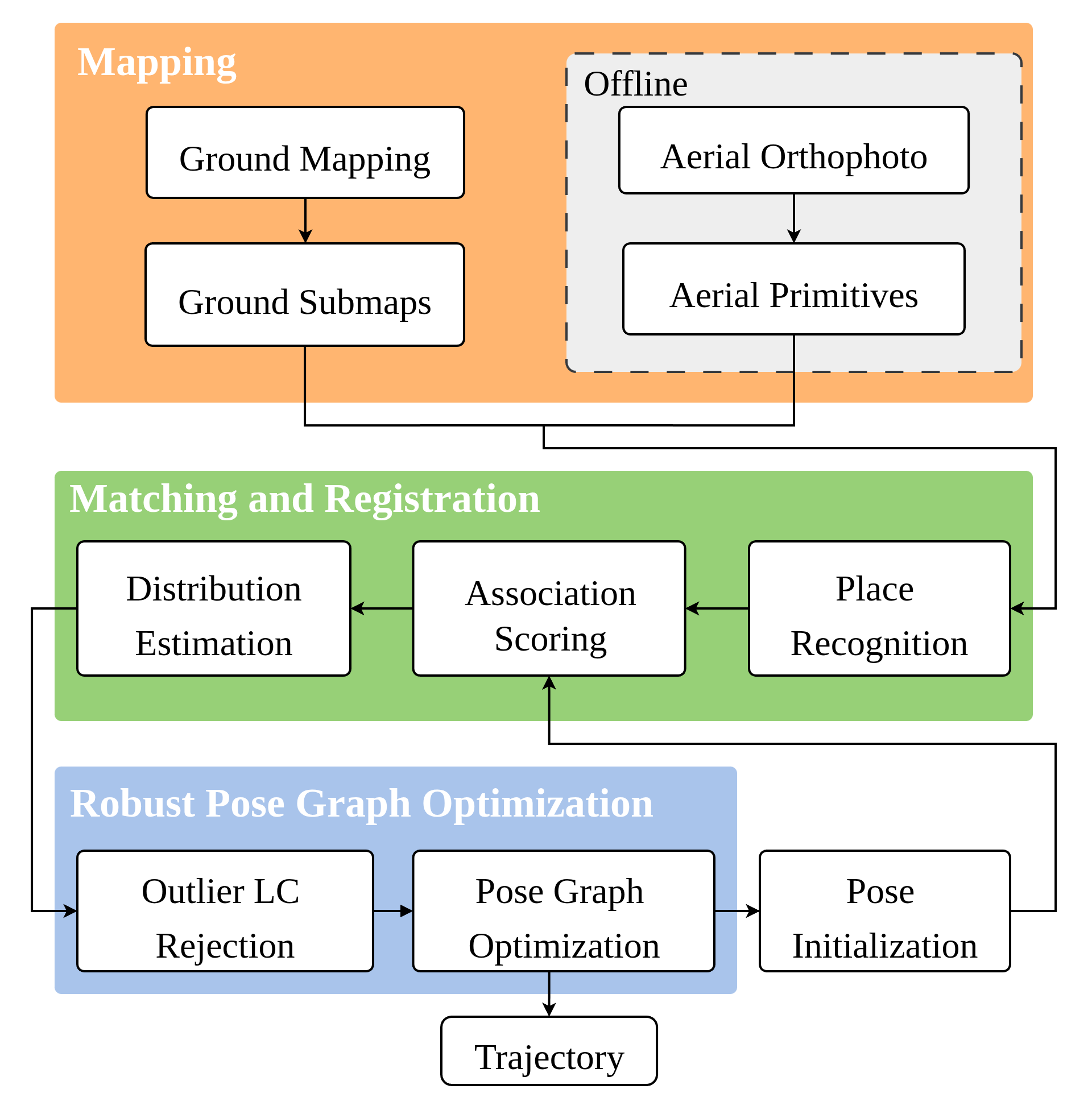}
    \caption{
    Our cross-view localization pipeline begins by extracting sparse primitives, grouped in aerial patches and ground submaps.
    The matching and registration module uses submaps and aerial patches to detect cross-view loop closures.
    Then, pose graph optimization is performed to estimate the ground trajectory.
    Once a rough trajectory is committed to, the resulting rotation can be used to guide the matching algorithm.
    We run mapping at a fixed 6 Hz and trigger submap creation after 20 new 3D segments are seen.
    }
    \label{fig:sys-diag}
    \vspace{-10pt}
\end{figure}

We propose a method for estimating a geo-stamped trajectory of a robot from aerial imagery using a single limited-field-of-view camera with depth measurements.
The fundamental concept we leverage is that high-level point and line features are useful as a sparse environment representation for matching features between aerial and ground views (see~\Cref{fig:intro-fig}).
A diagram of our cross-view mapping, matching, and trajectory estimation pipeline is shown in~\Cref{fig:sys-diag}, broken up into three submodules: Mapping (\Cref{sec:method-mapping}), Matching and Registration (\Cref{sec:method-outlier-rej,sec:method-matching}), and Robust Pose Graph Optimization (\Cref{sec:method-rpgo}).

In an offline mapping phase, sparse points and lines (i.e. {\it primitives}) with semantic descriptors are extracted from an aerial image.
Then, as a robot navigates in the area, it uses RGB-D camera input to create a map of primitives, which are grouped into \emph{submaps}.
Initially, the robot operates with no global location knowledge, so a place recognition module is used to recommend aerial patches for ground submap registration.
Our graph-theoretic data association algorithm is employed to match points and lines between the ground and aerial views, and these matches are used to produce a probability distribution of the transformation between the submap center and the associated aerial patch.
Likely pose hypotheses are passed as potential loop closures to our outlier rejection module, which eventually commits to a rough global localization estimate after sufficiently many consistent loop closures have been observed. 
Once a rough global location has been obtained, 
we use the pose estimate to determine geographically proximate aerial patches (thus skipping place recognition) and provide a prior on rotation in submap registration, enabling more accurate pose refinement. 

\subsection{Ground and Aerial Primitive Extraction}
\label{sec:method-mapping}

The central goal of our sparse primitive extraction approach is to find a set of sparse, high-level features represented in both the aerial and ground views.
We find that representing objects as points and region boundaries as lines (\eg, edge of a road or patch of grass) sparsely encodes important features observable from both views.

\begin{figure}[t]
    \centering
    \includegraphics[width=0.48\textwidth]{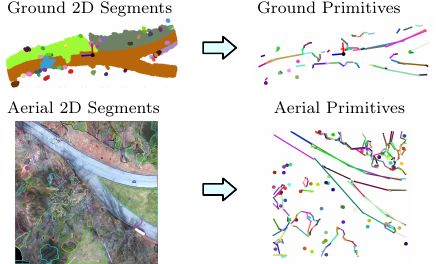}
    \caption{
    Example aerial and ground segment maps converted to point and line primitives. 
    Ground segment maps are built in 3D and then projected to 2D, while Segment Anything~\cite{kirillov2023segment} is used to directly create 2D segments from aerial images.
    Ground and aerial 2D segments are then converted to point and line primitives (right).
    }
    \label{fig:maps}
    \vspace{-0.4cm}
\end{figure}

To extract aerial points and lines, we first split an aerial reference image into overlapping aerial patches and run Segment Anything~\cite{kirillov2023segment} on each aerial patch.
Each segment with an area below a threshold is represented as a single point feature defined by the segment centroid.
Alpha shapes are then used to extract the border of each large segment, and fine-grained borders are fused together into coarse lines based on proximity and semantic and directional similarity.
Additionally, we run DINOv2~\cite{oquab2023dinov2} on each aerial patch to assign each primitive a DINOv2 semantic embedding.
Since the aerial imagery is typically obtained before a ground robot experiment, this computation can be done offline beforehand.

Ground metric-semantic primitives are created by first mapping segments as 3D point clouds with ROMAN~\cite{peterson2025roman}.
To enable incremental localization, newly observed 3D segments are stored in a queue, and the most recently seen segments are incorporated into a 3D submap when the queue is sufficiently large.
3D segments in each new submap are projected to 2D and converted to sparse points and lines following the same procedure as aerial primitive extraction.
The camera frustum of an RGB-D camera with a limited field of view (FOV) and range tends to introduce line artifacts; therefore, we mark segment points along the FOV boundaries and at the maximum depth range as \emph{border points}, and retain only lines that are sufficiently distant from these points.
We show example aerial and ground segment maps and the 
resulting primitives in~\Cref{fig:maps}.

\subsection{Graph-Theoretic Outlier Rejection}
\label{sec:method-outlier-rej}

As both the primitive matching and the robust pose graph optimization leverage similar outlier rejection technology, we introduce the common theoretical foundation here.
Given an outlier-polluted set of measurements, a consistency graph $\mathcal{G} = (\mathcal{N}, \mathcal{E})$ can be constructed, where nodes represent a measurement, and weighted edges exist between measurements that are deemed consistent.
An edge weight $e_{i,j}\in[0,1]$ encodes the consistency between measurements $i$ and $j$, where an edge weight of 0 denotes that a pair of measurements are inconsistent.
The graph $\mathcal{G}$ can be represented with an \emph{affinity matrix}, $\M{M} \in [0,1]^{|\mathcal{N}| \times |\mathcal{N}|}$, where each element $\M{M}(i,j) = e_{i,j},\, i,j \in  \mathcal{N}$.

Peterson et al.~\cite{peterson2025roman} showed that multiple measures of consistency can be effectively incorporated by taking their geometric mean, where for any $n \in \mathbb{N}$ and $x_1, \dots, x_n \in \mathbb{R}$, the geometric mean $\GM(x_1,\dots,x_n) \triangleq \left(\Pi_{i=1}^n x_i\right)^{\frac{1}{n}}$.
We combine pairwise consistency and individual measurement quality by letting
\begin{equation}
    \M{M}(i,j) = \GM \left(
       s(i,j), q(i), q(j)
    \right)
    \label{eq:affinity-score}
\end{equation}
where $s(i,j) \in [0, 1]$ is a scoring function measuring pairwise consistency, and $q(i) \in [0, 1]$ is a scoring function for the quality of a single input measurement.
For pairwise measurements, we compute a consistency score $f \in [0,1]$ by scoring a difference term $d$ with a noise term $\sigma$,
\begin{equation}
    f(d,\sigma) = \begin{cases} \exp\left(
        \frac{
            -d^2
        }{
            2\sigma^2
        } 
    \right) &\,|d| < \sigma \\
    0 &\,\text{otherwise.}
    \end{cases}
\end{equation}
Specific uses of $f$ are discussed in~\Cref{sec:method-matching,sec:method-rpgo}.

Once a consistency graph is constructed, the goal of outlier rejection becomes finding a set of mutually consistent nodes within the graph.
We leverage the insights of Lusk~et~al.~\cite{lusk2024clipper}, who further proposed that inliers tend to be found in the densest set of mutually consistent nodes.
Then, a robust set of inlier measurements can be found by solving
\begin{equation}
    \begin{split}
        \underset{\ve{u} \in \{0, 1\}^n}{\max}  &\frac{\ve{u}^\top \M{M} \ve{u}}{\ve{u}^\top \ve{u}}. \\
        \text{subject to} \quad &\ve{u}_i \ve{u}_j = 0 \; \text{if}\;  \M{M}_{i,j} = 0, \; \forall {i,j}.
    \end{split}
    \label{eq:clipper}
\end{equation}
Furthermore, Jia et al.~\cite{jia2025distribution} showed that approximate Bayesian inference techniques can be applied to this problem, yielding an approximate probability distribution over the possible sets of inlier measurements.
This distribution is represented as a set of particles, which can be clustered into nearby solutions to obtain a smaller set of likely hypotheses.
By maintaining multiple candidate solutions during front-end inlier selection, we are able to incorporate these probability-weighted hypotheses into a back-end solver, giving it more complete information and enabling the discovery of good solutions in challenging, high-outlier regimes.

\subsection{Metric-Semantic Primitive Matching}
\label{sec:method-matching}

Constructing initial probabilistic loop closure hypotheses begins with the creation of a ground submap.
Once created, we select a subset of aerial patches that will be used for primitive matching and registration. 
We use AnyLoc~\cite{keetha2023anyloc} to create global descriptors for each aerial patch in the offline aerial map creation phase.
Similarly, AnyLoc is run at regular intervals on ground robot camera frames, and each ground submap receives a small set of global descriptors from frames contained in that submap.
We run primitive matching and registration on the top $K_\text{global}$ patches with the highest global descriptor similarity to any of the ground submap's set of global descriptors.

In the matching module, points and lines from an aerial patch and a ground submap are used to construct a set of putative associations based on the $K_\text{prim}$ most semantically similar primitives.
Each association $a_i = (\objaerial_{u_i}, \objground_{v_i})$ is between a pair of points or between a pair of lines, with superscripts $a$ and $g$ denoting aerial and ground primitives respectively, and $u_i$ and $v_i$ representing the aerial and ground primitive indices belonging to the $i$-th association.
Given these putative associations, correspondence consistency is represented with a data association affinity matrix, $\M{M}_\text{DA}$, where $\M{M}_\text{DA}(i,j) \in [0, 1]$ follows the general scoring formula in~\Cref{eq:affinity-score} with custom scoring functions $s_\text{DA}$ and $q_\text{DA}$.
Then, the outlier rejection described in~\Cref{sec:method-outlier-rej} is used to identify a dense subset of mutually consistent associations. 

Importantly, we propose a novel scoring function for point-to-point and line-to-line associations that differentiates between infinite lines and bounded rays and line segments. 
Existing approaches for line associations use the affine Grassmannian metric~\cite{lusk2022graffmatch}, which treats lines as affine subspaces.
This is limiting for two reasons.
First, since all non-parallel infinite 2D lines intersect, the Grassmannian metric between lines reduces to only the angle between two lines, which does not discriminate enough between correct and spurious matches in 2D. 
Second, it is not clear how to incorporate known rotation or translation priors into the Grassmannian metric, which we show are helpful when a coarse rotation or translation estimate is available.

\begin{figure}[t]
    \centering
    \includegraphics[width=0.44\textwidth]{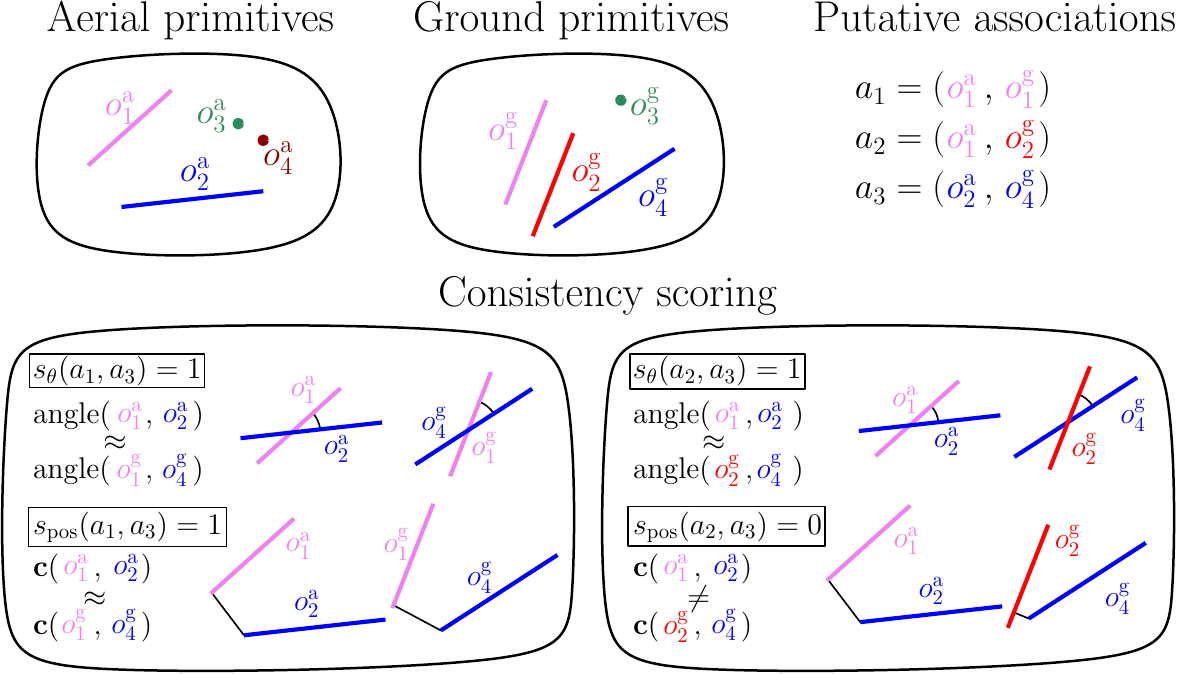}
    \caption{
    Point and line association consistency scoring. 
    The ground truth associations are indicated by color.
    On the bottom left, $\objaerial_1$ and $\objaerial_2$ are correctly matched with $\objground_1$ and $\objground_4$.
    This match is validated via a distance and angle check.
    On the right, $\objaerial_1$ and $\objaerial_2$ are matched with $\objground_2$ and $\objground_4$. 
    Although the two angles are consistent, the shortest distance between the two lines is inconsistent, so these association pairs are deemed inconsistent.
    }
    \label{fig:point-line-match}
    \vspace{-0.4cm}
\end{figure}

We instead propose a scoring metric that is general to both infinite and bounded lines and allows incorporation of a known rotation constraint.
We define 
\begin{equation}
    s_\text{DA}(a_i, a_j) = 
    \begin{cases}
        \GM(s_\theta(a_i, a_j), s_\text{t}(a_i, a_j)), &\text{is\_line}(a_i, a_j) \\
        s_\text{t}(a_i, a_j), &\text{otherwise}.
    \end{cases}
\end{equation}
Scoring functions $s_\theta$ and $s_\text{t}$ are illustrated between two primitive sets in~\Cref{fig:point-line-match}.
We define the directional scoring function $s_\theta$ to be 
\begin{equation}
    s_\theta(a_i, a_j) = f \left(
    \text{angle}(\objaerial_{u_i}, \objaerial_{u_j}) - 
    \text{angle}(\objground_{v_i}, \objground_{v_j})
    ,\sigma_\theta
    \right),
\end{equation}
where $\text{angle}(o_{u_i}, o_{u_j}) = \cos^{-1}\left(\left|\ve{d}(o_{u_i})^\top \ve{d}(o_{u_j})\right|\right)$ measures the angle between two line directions, with $\ve{d}(o_{u_i})$ being the normalized direction vector of a line $o_{u_i}$.
By taking the absolute value, we ensure that we take the smaller of the two angles formed by two lines.
Thus, $s_\theta(a_i, a_j)$ measures the difference between the two angles formed by the pair of lines with a noise term $\sigma_\theta$.
The two angles should be invariant between modalities if the two associations $a_i$ and $a_j$ are consistent.

The translational invariant $s_\text{t}(a_i, a_j)$ scores the consistency between pairs of associations of any primitive type.
Let $\ve{c}(o_{u_i}, o_{u_j}) = \Tilde{\ve{p}}(o_{u_i}, o_{u_j}) - \Tilde{\ve{p}}(o_{u_j}, o_{u_i})$,
where $\tilde{\ve{p}}(o_{u_i}, o_{u_j}) \in \R^3$ returns the point on primitive $o_{u_i}$ that is nearest to object $o_{u_j}$.
Thus, $\ve{c}$ defines a vector between the two closest points on objects $o_{u_i}$ and $o_{u_j}$.
Then, we define the translational invariant as
\begin{equation}
\label{eq:point-invariant}
    s_\text{t}(a_i, a_j) = f\left(
        \|\ve{c}(o_{u_i}^a, o_{u_j}^a)\| - \|\ve{c}(o_{v_i}^g, o_{v_j}^g)\|, \sigma_\text{t}
    \right).
\end{equation}
The key insight here is that the distance between the two closest points on two primitives is invariant to rotation and translation, so we can use this distance as a consistency check on pairs of associations.
Furthermore, if a prior on rotation exists (\eg, from a rough initial pose after the system has globally localized), $s_\text{t}$ and $s_\theta$ can naturally incorporate the known rotation by splitting $s_\text{t}$ into $x$ and $y$ components and evaluating absolute angles with $s_\theta$.
In this way, high-scoring matches that are not consistent with a known rotation direction can be rejected.
Finally, following the findings of~\cite{peterson2025roman}, we incorporate the semantics into the singular quality score $q_\text{DA}$ by taking the scaled cosine similarity of the associated primitives' DINOv2~\cite{oquab2023dinov2} descriptors.
Using $s_\text{DA}$ and $q_\text{DA}$, the data association affinity matrix $\M{M}_\text{DA}$ is computed and~\cite{jia2025distribution} is used to solve for a distribution of inlier data association sets represented as particles.

We use each particle's association vector to compute
a 2D transformation between the center of the aerial patch and the center of the ground submap using a modification of Arun's method~\cite{arun1987least}. 
Line directions are treated equivalently to centroid-adjusted points when constructing the cross-covariance matrix, ensuring that the rotation aligns both points and line directions. Translation is estimated by jointly minimizing point and Plücker line coordinate errors via weighted least squares. To resolve line direction ambiguities, a minimal non-degenerate subset of associated points and/or lines is selected, and all possible line direction combinations are enumerated, with a transform computed for each. The transform yielding the smallest error across all primitives is used to fix line directions, after which a refined transform is computed using all inlier associations.
We then cluster the transformations estimated from all association particles in pose space. 
The top $L$ largest clusters by cardinality are selected as pose estimate hypotheses.
This procedure produces a potentially multi-modal pose distribution, which is provided to the robust pose graph optimization module.

\subsection{Robust Pose Graph Optimization}
\label{sec:method-rpgo}

\begin{figure}[t]
    \centering
    \includegraphics[width=0.48\textwidth]{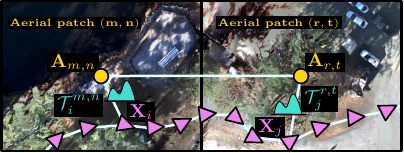}
    \caption{
    Cross-view pose graph.
    Aerial patches are connected via known translations, and pose variables are connected via odometry estimates. 
    Ground poses $\M{X}_i$ and $\M{X}_j$ estimate a distribution of potential poses in aerial patches, $\M{A}_{m,n}$ and $\M{A}_{r,t}$ respectively.
    Individual hypotheses in $\lcdist$ and $\mathcal{T}^{r,t}_{j}$ can be evaluated for consistency by comparing the chained odometry transform $\mathbf{T}_\text{odom}$ and the chained loop closure transform $\mathbf{T}_\text{LC}$.
    }
    \label{fig:pose-graph}
    \vspace{-0.4cm}
\end{figure}

We formulate cross-view trajectory estimation as a robust pose graph optimization (RPGO) problem where the graph consists of ground nodes $\Tground{k} \in \text{SE}(3)$, $k = 1, \dots, K$, representing the $K$ ground poses, and aerial nodes $\Taerial{m,n} \in \text{SE}(3)$, $m = 1, \dots, M$, $n = 1, \dots, N$, representing an aerial image divided into $M \times N$ patches.
Since aerial patches are already in a common global frame, we only optimize the ground poses.
We write the cross-view registration probability distribution obtained from our registration module as $\lcdist$ where $(m,n)$ denotes the aerial patch and $k$ denotes the ground pose index.
Concretely, we represent this distribution as a set of $L$ hypotheses, such that $\lcdist = \left\{ \left( \lc, p^{m,n}_{k,\ell}\right) \right\}_{\ell=1}^L$,
where $\lc$ represents a single hypothesis transformation corresponding to a probability $p^{m,n}_{k,\ell}$.
We approximate each normalized probability $p^{m,n}_{k,\ell}$ as the number of particles that were part of the corresponding cluster $\lc$ divided by the total number of particles.

In practice, the collection of all submap-to-aerial-patch probability distributions contains many incorrect individual hypotheses, which stems from two factors: first, place recognition in cross-view regimes is difficult, requiring registration attempts on a large number of aerial patches, and second, cross-view registration tends to be highly multi-modal.
Inspired by~\cite{mangelson2018pairwise}, we approach this problem with the same unified outlier rejection mechanism outlined in~\Cref{sec:method-outlier-rej}.
In this scenario each element of the affinity matrix $\M{M}_\text{RPGO}$ represents consistency between a pair of loop closures rather than a pair of associations.
Each $\M{M}_\text{RPGO}(i,j)$ considers two discrete loop closure measurements, $\M{T}^{m,n}_i$ and $\M{T}^{r,t}_j$, where we drop the hypothesis subscript for simplicity. 
Then the two measurements, $\M{T}_\text{odom} = \M{X}_i^{-1} \M{X}_j$ and $\M{T}_\text{LC} = \left(\M{A}_{m,n}\M{T}^{m,n}_i\right)^{-1}  \M{A}_{r,t} \M{T}^{r,t}_j$, should be similar to each other if the two loop closures are consistent.
We use the transforms to construct the pairwise score,
\begin{multline}
    s_\text{RPGO}\left(\M{T}^{m,n}_i, \M{T}^{r,t}_j\right) = 
    \text{GM} \left(
        f\left(
            \psi(\M{T}_\text{odom}^{-1} \M{T}_\text{LC}), \sigma_\psi(i,j)
        \right), \right.
        \\ \left.
        f\left(
            \| \ve{t}_\text{odom} - \ve{t}_\text{LC} \|, \sigma_t(i,j)
        \right)
    \right)
\end{multline}
where $\psi$ extracts the 2D yaw from a transformation, $\ve{t}_\text{odom}$ and $\ve{t}_\text{LC}$ represent the translational components of the corresponding transformations, and $\sigma_\psi$ and $\sigma_t$ are rotation and translation noise terms. 
While maximum clique approaches are a common way of rejecting outliers in pose graph optimization~\cite{mangelson2018pairwise},
our key contribution is defining $\sigma_\psi$ and $\sigma_t$ as functions that scale linearly with path length between the two robot poses. 
This models the marginal uncertainty between two poses connected by odometry measurements, without the heavy computation of calculating that uncertainty for every pair of ground submaps.
Without scaling the uncertainty terms, outlier rejection is overly conservative and cannot account for the drift that occurs from robot odometry. 
Other works have addressed this with a second solver using graduated non-convexity after an initial frame alignment~\cite{tian2022kimera}; however, our single step process is more computationally reasonable and avoids cascaded failures.

Furthermore, we can naturally integrate the different estimated probabilities of different individual hypotheses via the quality score, which we define as 
$
q_\text{RPGO}\left(\lc\right) = p^{m,n}_{k,\ell} / \max_{\tilde{\ell}\in 1..L} p^{m,n}_{k,\tilde{\ell}}.
$
Then, once $\M{M}_\text{RPGO}$ is computed, we use CLIPPER~\cite{lusk2024clipper} to solve for a set of inlier loop closures.
Outlier loop closures are removed from the pose graph and pose graph optimization is performed to estimate the robot trajectory in the global frame.

\section{Experiments}
\label{sec:experiments}

\begin{figure*}[t]
    \centering
    
    \includegraphics[width=0.94\textwidth,height=0.7\textheight,keepaspectratio]{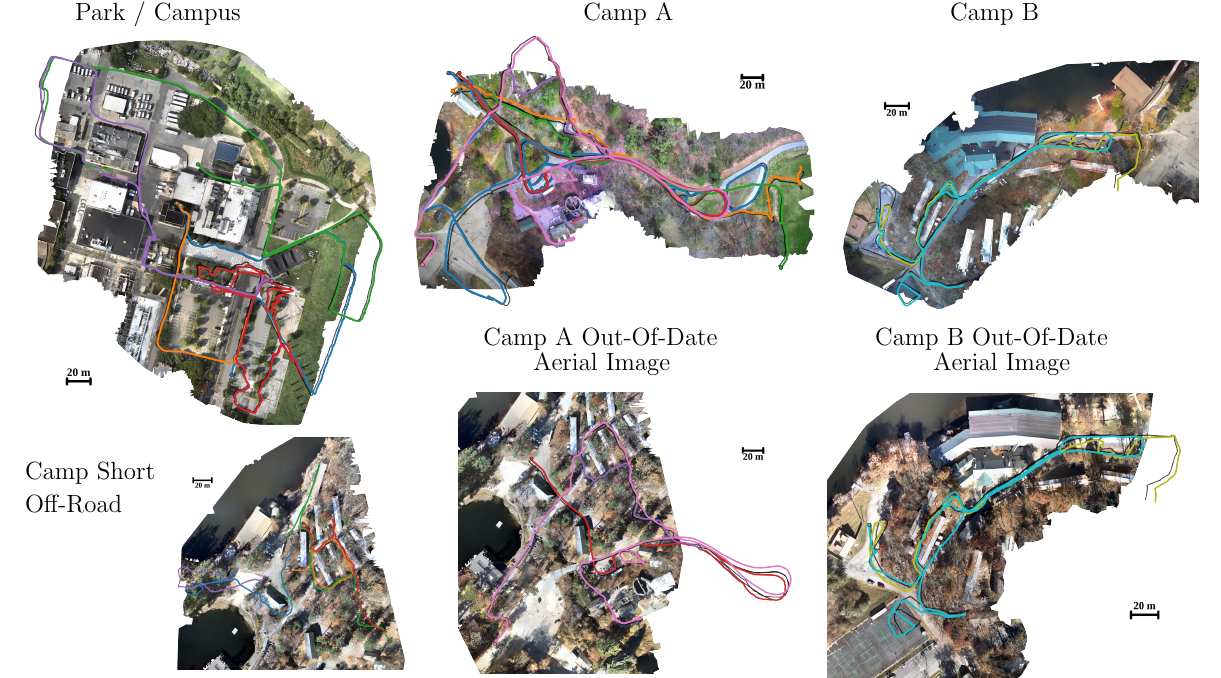}

    \caption{
    Top row and bottom right two images: self-collected aerial images and the final optimized trajectory of each sequence shown in a unique color (corresponding to sequence colors in~\Cref{tab:ate-park-campus,tab:ate-camp}) with the ground truth drawn in black.
    Bottom left: trajectories used for the off-road submap registration experiment.
    }
    \label{fig:results-main}
    \vspace{-10pt}
\end{figure*}

We evaluate the full Meridian pipeline for robot cross-view trajectory estimation from an aerial image, and additionally perform isolated ground-to-aerial submap registration experiments.
Represented regions span diverse urban and natural environments, testing generalization without environment-specific training or fine-tuning.

\textbf{Datasets}
First, we test Meridian using the KITTI odometry dataset~\cite{geiger2013vision}, showcasing a large-scale urban environment.
We use KISS-ICP~\cite{vizzo2023ral} for odometry and LiDAR-projected depth maps for mapping.
Aerial imagery is obtained from Vexcel.
Next, we evaluate Meridian on a self-collected Park/Campus Dataset with five robot trajectories totaling \SI{3.1}{km} (\Cref{fig:results-main}). 
The traversals cover parkland, sidewalks, parking lots, and narrow alleyways, extending evaluation beyond autonomous-driving settings.
We use an aerial orthomosaic from drone imagery collected in November 2024. 
RGB-D ground data was collected a year later (November 2025) on a heterogeneous team of two Clearpath Husky UGVs, two Boston Dynamics Spots, and an Agile-X Scout Mini with ZED 2i cameras.
The Husky and Scout UGVs use DLIO~\cite{chen2022direct} and the Spots use onboard visio-kinematic odometry.
Ground-truth pose was obtained by fusing GPS and odometry.

Finally, we evaluate Meridian on our self-collected Camp A and Camp B datasets with 10 trajectories totaling \SI{6.3}{km}. 
Most of the traversal is constrained to roads due to traversability limitations, but roughly one fifth covers rocky off-road areas, fields, and trails, enabling cross-view localization evaluation in previously untested environments.
We evaluate all 10 sequences on same-week aerial imagery and five sequences (which maintain sufficient overlap with the different geographic footprint of the older aerial imagery) on 1.5-year-old aerial imagery. 
The older imagery contains seasonal changes, moved objects, and partial coverage differences. 
Ground data was collected in April 2026 with the same sensor suite as the Park/Campus dataset, and ground-truth poses are obtained from GPS and odometry with manual correction for GPS-degraded sequences A-5 and A-7.

\textbf{Implementation Details}
We use \SI{60}{m}$\times$\SI{60}{m} aerial patches with 50\% overlap and ground submaps of 120 3D segments.
We set $K_\text{global}=25$, $K_\text{prim}=20$, $\sigma_t=$\SI{2}{m}, $\sigma_\theta=$\SI{20}{deg}, and $L=25$.
Rough global localization is accepted after $\geq$ 6 loop closures are accepted as inliers and the inlier set makes up at least 50\% of all ground submaps. 

\subsection{Full Pipeline Experiments}
\label{sec:exp-full}
We begin by running the Meridian cross-view localization pipeline on the KITTI odometry sequences 00, 02, and 09, following~\cite{miller2021any}.
We report RMS absolute trajectory error (ATE) in~\Cref{tab:kitti-ate}, computed between OXTS ground truth and Meridian's estimated trajectory.
The reported ``incremental'' ATE reflects the instantaneous pose estimate error at each timestamp after Meridian has committed to a global localization, and we additionally report the RMSE of the final optimized trajectory.
We include OpenStreetMap-based methods for reference, though they require mapped roads and are therefore limited to urban settings. 
In contrast, Meridian requires only overhead imagery and matches or outperforms prior aerial-image-based methods in both incremental and final ATE.

\begin{table}[t]
      \scriptsize
      \centering
      \caption{KITTI ATE (m)}  
      \setlength{\tabcolsep}{8pt}
      \begin{tabular}{c l c c c}
          \toprule
          & \textbf{Method}
          & \textbf{Seq.\ 00} & \textbf{Seq.\ 02} & \textbf{Seq.\ 09} \\
          \midrule

          \multirow{3}{*}{\rotatebox{90}{\textbf{OSM}}}
          & {OSM-SLAM}~\cite{frosi2023osm}                  & 3.8          & \textbf{1.3} & --            \\
          & {Przewodowski}~\cite{przewodowski2024global}     & 2.8          & 3.1          & 3.6           \\
          & {Li}~\cite{li2026odometry}*                      & \textbf{1.6} & failed         & \textbf{1.1}  \\
          \midrule

          \multirow{4}{*}{\rotatebox{90}{\textbf{Aerial}}}
          & {Miller}~\cite{miller2021any}                    & 2.0          & 9.1          & 7.2           \\
          & {Ankenbauer}~\cite{ankenbauer2023global}         & 3.5          & 4.4          & 3.0           \\
          & {Meridian (incremental)}                         & 2.0 & \textbf{3.3} & {2.4}  \\
          & {Meridian (final trajectory)}                    & \textbf{1.5}          & 4.2          & \textbf{2.1}            \\
          \bottomrule
      \end{tabular}

      \vspace{2pt}
      \footnotesize
      \parbox{0.9\linewidth}{
      * Did not report whether error is reported as RMSE or mean or whether error is absolute in a global position sense or whether~\cite{horn1987closed} was used to provide a
  least-squares alignment.
      }

      \label{tab:kitti-ate}
      \vspace{-.2cm}
  \end{table}

\begin{table}[t]
    \scriptsize
    \centering
    \caption{Trajectory Estimation Error on Park/Campus Dataset}
    \setlength{\tabcolsep}{5pt}
    \begin{tabular}{c c c c c}
        \toprule
        \textbf{Sequence} &
        \begin{tabular}{c}\textbf{Robot}\\\textbf{Type}\end{tabular} &
        \begin{tabular}{c}\textbf{Path}\\\textbf{Length (m)}\end{tabular} &
        \begin{tabular}{c}\textbf{Incremental}\\\textbf{ATE (m)}\end{tabular} &
        \begin{tabular}{c}\textbf{Final}\\\textbf{ATE (m)}\end{tabular} \\
        \midrule
            \textcolor{tabblue}{\textbf{1}}   & Spot   & 504  & 3.22 & 1.77 \\
            \textcolor{taborange}{\textbf{2}}    & Husky  & 239  & 1.07 & 1.22 \\
            \textcolor{tabgreen}{\textbf{3}} & Spot   & 951  & 2.52 & 2.25 \\
            \textcolor{tabred}{\textbf{4}}    & Husky  & 694  & 2.2 & 2.38 \\
            \textcolor{tabpurple}{\textbf{5}}   & Scout  & 734  & 1.56 & 1.07 \\
        \midrule
        \textbf{Mean / Std} & -- & -- & $2.11 \pm 0.75$ & $1.74 \pm 0.53$ \\

        \bottomrule
        \vspace{-0.6cm}
    \end{tabular}
    \label{tab:ate-park-campus}
\end{table}

  \begin{table}[t]
      \scriptsize
      \centering
      \caption{Trajectory Estimation Error on Camp Dataset}
      \setlength{\tabcolsep}{5pt}
      \begin{tabular}{c c c c c c c}
          \toprule
          \multirow{4}{*}{\textbf{Seq.}} &
          \multirow{4}{*}{\begin{tabular}{c}\textbf{Robot}\\\textbf{Type}\end{tabular}} &
          \multirow{4}{*}{\begin{tabular}{c}\textbf{Path}\\\textbf{Len. (m)}\end{tabular}} &
          \multicolumn{4}{c}{\textbf{ATE (m)}} \\
          \cmidrule(lr){4-7}
          & & & \multicolumn{2}{c}{\textbf{Up-To-Date Aerial}} & \multicolumn{2}{c}{\textbf{Outdated Aerial}} \\
          \cmidrule(lr){4-5} \cmidrule(lr){6-7}
          & & & \textbf{Inc.} & \textbf{Final} & \textbf{Inc.} & \textbf{Final}
          \\
          \midrule
              \textcolor{tabblue}{\textbf{A-1}}  & Spot     & 1426 & 4.50 & 2.61 & & \\
              \textcolor{taborange}{\textbf{A-2}}  & Husky    & 467  & 4.48 & 3.15 & & \\
              \textcolor{tabgreen}{\textbf{A-3}}  & Husky    & 409  & 2.51 & 2.16 & & \\
              \textcolor{tabred}{\textbf{A-4}}  & Spot     & 458  & 1.53 & 1.38 & 3.00 & 2.27 \\
              \textcolor{tabpurple}{\textbf{A-5}}  & Spot     & 211  & 1.42 & 2.02 & 1.79 & 1.6 \\
              \textcolor{tabbrown}{\textbf{A-6}}  & Spot     & 506  & 2.04 & 2.78 & & \\
              \textcolor{tabpink}{\textbf{A-7}}  & Husky    & 996  & 3.66 & 2.09 & 7.25 & 4.07 \\
              \textcolor{tabgray}{\textbf{B-1}}  & Spot     & 597  & 4.72 & 2.94 & & \\
              \textcolor{tabolive}{\textbf{B-2}}  & Spot     & 448  & 2.60 & 2.27 & 4.12 & 3.80 \\
              \textcolor{tabcyan}{\textbf{B-3}}  & Spot     & 755  & 2.87 & 2.23 & 4.34 & 2.84 \\
          \midrule
          \textbf{Mean} & -- & -- & 3.03 & 2.38 & 4.10 & 2.91 \\
          \textbf{Std}  & -- & -- & 1.18 & 0.47 & 1.82 & 0.93 \\
          \bottomrule
          \vspace{-.5cm}
      \end{tabular}
      \label{tab:ate-camp}
  \end{table}

Furthermore, Meridian generalizes well to environments beyond urban driving scenes, demonstrated by performance on the Park/Campus and Camp sequences in~\Cref{tab:ate-park-campus,tab:ate-camp}.
These environments include featureless landscapes and repetitive geometries including sidewalks, parking lots, rural roads, fields, forested areas, and off-road terrain.
Across the Park/Campus and Camp sequences, Meridian achieves an average incremental ATE of \SI{3.07}{m} and final ATE of \SI{2.35}{m} over 9 km of unique traversal. 
Using 1.5-year-old aerial imagery, Meridian exhibits only modest degradation despite seasonal changes, moved objects, and partial map coverage.

\subsection{Submap Registration Experiments}
\label{sec:exp-submap-reg}

\begin{table}[t]
    \centering
    \scriptsize
    \caption{Registration Success Rate (Percentage \textless 5 m, \textless 5 deg)}
    \label{tab:registration-success}
    \renewcommand{\arraystretch}{1.15}
    \setlength{\tabcolsep}{2pt}

    \newcommand{\datasetlabel}[1]{%
        \rotatebox[origin=c]{90}{\makecell[c]{#1}}%
    }

    \resizebox{\columnwidth}{!}{%
    \begin{tabular}{c l c c c c c c c c}
        \toprule
        Dataset
        & Method
        & \multicolumn{4}{c}{Known Rotation}
        & \multicolumn{4}{c}{Unknown Rotation} \\
        \cmidrule(lr){3-6}
        \cmidrule(lr){7-10}
        &
        & Top-$1$ & Top-$5$ & Top-$10$ & Top-$25$
        & Top-$1$ & Top-$5$ & Top-$10$ & Top-$25$ \\
        \midrule

        \multirow{3}{*}{\datasetlabel{KITTI\\Seq. 00}}
        & Meridian
        & 38.3 & 71.3 & 82.4 & 92.7
        & \textbf{23.0} & \textbf{44.1} & \textbf{53.6} & \textbf{62.8} \\
        & Meridian w/o Lines
        & 41.8 & 70.1 & 76.6 & 80.1
        & 22.2 & 36.0 & 43.7 & 56.3 \\
        & FG$^2$
        & \textbf{77.0} & \textbf{95.4} & \textbf{96.6} & \textbf{98.9}
        & -- & -- & -- & -- \\
        \midrule

        \multirow{3}{*}{\datasetlabel{Park/\\Campus}}
        & Meridian
        & \textbf{53.5} & \textbf{80.7} & \textbf{89.5} & \textbf{92.1}
        & \textbf{39.5} & \textbf{61.4} & \textbf{66.7} & \textbf{73.7} \\
        & Meridian w/o Lines
        & 43.9 & 71.1 & 77.2 & 84.2
        & 29.8 & 38.6 & 45.6 & 59.6 \\
        & FG$^2$
        & 21.1 & 44.7 & 57.9 & 78.9
        & -- & -- & -- & -- \\
        \midrule

        \multirow{3}{*}{\datasetlabel{Camp\\All}}
        & Meridian
        & \textbf{33.7} & \textbf{61.1} & \textbf{69.3} & \textbf{79.1}
        & \textbf{15.7} & \textbf{27.5} & \textbf{36.3} & \textbf{46.7} \\
        & Meridian w/o Lines
        & 28.8 & 50.7 & 55.9 & 61.4
        & 8.8 & 17.3 & 23.9 & 34.0 \\
        & FG$^2$
        & 22.5 & 48.0 & 62.7 & 75.5
        & -- & -- & -- & -- \\
        \midrule

        \multirow{3}{*}{\datasetlabel{Camp\\Off-Road}}
        & Meridian
        & \textbf{30.3} & \textbf{53.0} & \textbf{62.1} & \textbf{72.7}
        & \textbf{9.1} & 13.6 & \textbf{22.7} & \textbf{36.4} \\
        & Meridian w/o Lines
        & 22.7 & 45.5 & 51.5 & 65.2
        & 4.5 & \textbf{16.7} & 18.2 & 34.8 \\
        & FG$^2$
        & 13.6 & 27.3 & 36.4 & 45.5
        & -- & -- & -- & -- \\

        \bottomrule
    \end{tabular}%
    }
\end{table}

\begin{table}[h!]
\scriptsize
\centering
\caption{RPGO Experiment: Camp A ATE (m) (Inc. / Final)}
\setlength{\tabcolsep}{2.5pt}
\begin{tabular}{l c c c c c c c}
\toprule
 & \textbf{A-1} & \textbf{A-2} & \textbf{A-3} & \textbf{A-4} &
 \textbf{A-5} & \textbf{A-6} & \textbf{A-7} \\
\midrule
Meridian &
4.5 / 2.6 &
4.5 / 3.2 &
2.5 / 2.2 &
1.5 / 1.4 &
1.4 / 2.0 &
2.0 / 2.8 &
3.7 / 2.1
\\

w/o scaling &
- / 17.2 &
- / 20.1 &
- / 95.5 &
- / 2.5 &
- / 1.7 &
- / 7.2 &
- / 5.0

\\
\bottomrule
\end{tabular}
\label{tab:rpgo-ablation}
\vspace{-0.3cm}
\end{table}

We isolate the matching and registration module by splitting robot trajectories into submaps created every \SI{40}{m} and evaluate registration success on each ground submap in a \SI{60}{m}$\times$\SI{60}{m} aerial patch. 
We report the top-$L$ success rate for $L\in\{1,5,10,25\}$, where a hypothesis is successful if it is within \SI{5}{m} translation and \SI{5}{\degree} rotation error.
We highlight performance on off-road terrain by registering a subset of off-road submaps to the 1.5-year-old overhead imagery, which contains more of the off-road regions.

We compare Meridian to a variant, ``Meridian w/o Lines'', which only uses points for registration.
Additionally, we compare against \FGsq~\cite{xia2025fg}, a learning-based cross-view geo-localization method trained on KITTI data.
Since \FGsq~operates on a single ground and aerial image pair, we run it on each ground image in a Meridian submap and cluster its predicted poses for top-$L$ evaluation. 
We evaluate Meridian variants using both known and unknown rotation, but only report \FGsq~in a known-rotation setting since \FGsq~requires 360-degree imagery for unknown-rotation inference.

Although \FGsq~outperforms Meridian on KITTI submaps, we find that Meridian generalizes better across all environments and consistently outperforms \FGsq~on our self-collected data.
Furthermore, we find that our method for matching both points and lines results in higher registration success rate, especially in the Camp All and Park/Campus submaps, where region boundaries yield abundant line features.

\subsection{Loop Closure Outlier Rejection Experiments}

Finally, in~\Cref{tab:rpgo-ablation} we compare Meridian against a variant of our method that does not scale uncertainty with distance traveled in loop closure rejection (``w/o scaling'').
Without uncertainty scaling, incremental solutions are never accepted due to excessive rejection of loop closures.
As a result, the final trajectory errors are generally much higher as well, indicating that this variant is overly conservative.

\section{Conclusion}
\label{sec:conclusion}

We have demonstrated our cross-view localization pipeline across a range of urban and natural environments, and shown 
that metric-semantic primitives can be reliably associated across viewpoints to register ground robot maps with overhead aerial imagery. A limitation of the proposed work is a low recall rate from place recognition, requiring the algorithm to run matching on many non-overlapping ground submap and aerial patch pairs.
While these can be rejected downstream by our robust pose graph optimization, scalability would be improved with higher cross-view place recognition recall.
Furthermore, we have observed that DINOv2~\cite{oquab2023dinov2} feature similarity tends to be less discriminative across aerial and ground views than between two ground views.
Future work includes investigating methods to bridge the semantic differences between ground and aerial views.

\section*{Acknowledgment}

Portions of the code were developed using Claude Code.

\bibliographystyle{IEEEtran}
\bibliography{references}

@article{zhao2023fast,
  title={Fast segment anything},
  author={Zhao, Xu and Ding, Wenchao and An, Yongqi and Du, Yinglong and Yu, Tao and Li, Min and Tang, Ming and Wang, Jinqiao},
  journal={arXiv preprint arXiv:2306.12156},
  year={2023}
}

@inproceedings{kirillov2023segment,
  title={Segment anything},
  author={Kirillov, Alexander and Mintun, Eric and Ravi, Nikhila and Mao, Hanzi and Rolland, Chloe and Gustafson, Laura and Xiao, Tete and Whitehead, Spencer and Berg, Alexander C and Lo, Wan-Yen and others},
  booktitle={Proceedings of the IEEE/CVF International Conference on Computer Vision},
  pages={4015--4026},
  year={2023}
}

@article{lusk2024clipper,
  title={{CLIPPER: Robust Data Association without an Initial Guess}},
  author={Lusk, Parker C and How, Jonathan P},
  journal={IEEE Robotics and Automation Letters},
  year={2024},
  publisher={IEEE}
}

@inproceedings{ankenbauer2023global,
  title={Global localization in unstructured environments using semantic object maps built from various viewpoints},
  author={Ankenbauer, Jacqueline and Lusk, Parker C and Thomas, Annika and How, Jonathan P},
  booktitle={2023 IEEE/RSJ international conference on intelligent robots and systems (IROS)},
  pages={1358--1365},
  year={2023},
  organization={IEEE}
}

@article{arun1987least,
  title={Least-squares fitting of two 3-D point sets},
  author={Arun, K Somani and Huang, Thomas S and Blostein, Steven D},
  journal={IEEE Transactions on pattern analysis and machine intelligence},
  number={5},
  pages={698--700},
  year={1987},
  publisher={IEEE}
}

@article{ransac,
  title={Random sample consensus: a paradigm for model fitting with applications to image analysis and automated cartography},
  author={Fischler, Martin A and Bolles, Robert C},
  journal={Communications of the ACM},
  volume={24},
  number={6},
  pages={381--395},
  year={1981},
  publisher={ACM New York, NY, USA}
}

@inproceedings{mangelson2018pairwise,
  title={Pairwise consistent measurement set maximization for robust multi-robot map merging},
  author={Mangelson, Joshua G and Dominic, Derrick and Eustice, Ryan M and Vasudevan, Ram},
  booktitle={2018 IEEE international conference on robotics and automation (ICRA)},
  pages={2916--2923},
  year={2018},
  organization={IEEE}
}

@article{tian2022kimera,
  title={Kimera-multi: Robust, distributed, dense metric-semantic slam for multi-robot systems},
  author={Tian, Yulun and Chang, Yun and Arias, Fernando Herrera and Nieto-Granda, Carlos and How, Jonathan P and Carlone, Luca},
  journal={IEEE Transactions on Robotics},
  volume={38},
  number={4},
  year={2022},
  publisher={IEEE}
}

@article{yang2020teaser,
  title={Teaser: Fast and certifiable point cloud registration},
  author={Yang, Heng and Shi, Jingnan and Carlone, Luca},
  journal={IEEE Transactions on Robotics},
  volume={37},
  number={2},
  pages={314--333},
  year={2020},
  publisher={IEEE}
}

@article{dube2018incremental,
  title={Incremental-segment-based localization in 3-d point clouds},
  author={Dub{\'e}, Renaud and Gollub, Mattia G and Sommer, Hannes and Gilitschenski, Igor and Siegwart, Roland and Cadena, Cesar and Nieto, Juan},
  journal={IEEE Robotics and Automation Letters},
  volume={3},
  number={3},
  pages={1832--1839},
  year={2018},
  publisher={IEEE}
}

@incollection{sh-p1-prelude,
  title        = {Part1 Prelude},
  author       = {Luca Carlone and Ayoung Kim and Timothy Barfoot and Daniel Cremers and Frank Dellaert},
  booktitle    = {{SLAM Handbook.} From Localization and Mapping to Spatial Intelligence},
  publisher    = {Cambridge University Press},
  editor       = {Luca Carlone and Ayoung Kim and Timothy Barfoot and Daniel Cremers and Frank Dellaert},
  year         = {2026}
}

@article{vizzo2023ral,
  author    = {Vizzo, Ignacio and Guadagnino, Tiziano and Mersch, Benedikt and Wiesmann, Louis and Behley, Jens and Stachniss, Cyrill},
  title     = {{KISS-ICP: In Defense of Point-to-Point ICP -- Simple, Accurate, and Robust Registration If Done the Right Way}},
  journal   = {IEEE Robotics and Automation Letters (RA-L)},
  pages     = {1029--1036},
  doi       = {10.1109/LRA.2023.3236571},
  volume    = {8},
  number    = {2},
  year      = {2023},
  codeurl   = {https://github.com/PRBonn/kiss-icp},
}

@inproceedings{peterson2025roman,
  title={{ROMAN: Open-Set Object Map Alignment for Robust View-Invariant Global Localization}},
  author={Peterson, Mason B and Jia, Yi Xuan and Tian, Yulun and Thomas, Annika and How, Jonathan P},
  booktitle={Robotics: Science and Systems (RSS)},
  year={2025}
}

@article{keetha2023anyloc,
  title={{AnyLoc: Towards universal visual place recognition}},
  author={Keetha, Nikhil and Mishra, Avneesh and Karhade, Jay and Jatavallabhula, Krishna Murthy and Scherer, Sebastian and Krishna, Madhava and Garg, Sourav},
  journal={IEEE Robotics and Automation Letters},
  volume={9},
  number={2},
  pages={1286--1293},
  year={2023},
  publisher={IEEE}
}

@article{durgam2024cross,
  title={Cross-view geo-localization: a survey},
  author={Durgam, Abhilash and Paheding, Sidike and Dhiman, Vikas and Devabhaktuni, Vijay},
  journal={IEEE Access},
  year={2024},
  publisher={IEEE}
}

@inproceedings{lin2013cross,
  title={Cross-view image geolocalization},
  author={Lin, Tsung-Yi and Belongie, Serge and Hays, James},
  booktitle={Proceedings of the IEEE Conference on Computer Vision and Pattern Recognition},
  pages={891--898},
  year={2013}
}

@inproceedings{xia2025fg,
  title={FG\^{} 2: Fine-Grained Cross-View Localization by Fine-Grained Feature Matching},
  author={Xia, Zimin and Alahi, Alexandre},
  booktitle={Proceedings of the Computer Vision and Pattern Recognition Conference},
  pages={6362--6372},
  year={2025}
}

@inproceedings{hu2018cvm,
  title={Cvm-net: Cross-view matching network for image-based ground-to-aerial geo-localization},
  author={Hu, Sixing and Feng, Mengdan and Nguyen, Rang MH and Lee, Gim Hee},
  booktitle={Proceedings of the IEEE Conference on Computer Vision and Pattern Recognition},
  pages={7258--7267},
  year={2018}
}

@article{jia2025distribution,
  title={Distribution Estimation for Global Data Association via Approximate Bayesian Inference},
  author={Jia, Yixuan and Peterson, Mason B and Li, Qingyuan and Tian, Yulun and How, Jonathan P},
  journal={arXiv preprint arXiv:2509.15565},
  year={2025}
}

@article{lusk2022graffmatch,
  title={{GraffMatch: Global matching of 3d lines and planes for wide baseline lidar registration}},
  author={Lusk, Parker C and Parikh, Devarth and How, Jonathan P},
  journal={IEEE Robotics and Automation Letters},
  volume={8},
  number={2},
  pages={632--639},
  year={2022},
  publisher={IEEE}
}

@article{oquab2023dinov2,
  title={Dinov2: Learning robust visual features without supervision},
  author={Oquab, Maxime and Darcet, Timoth{\'e}e and Moutakanni, Th{\'e}o and Vo, Huy and Szafraniec, Marc and Khalidov, Vasil and Fernandez, Pierre and Haziza, Daniel and Massa, Francisco and El-Nouby, Alaaeldin and others},
  journal={arXiv preprint arXiv:2304.07193},
  year={2023}
}

@article{miller2021any,
  title={Any way you look at it: Semantic crossview localization and mapping with lidar},
  author={Miller, Ian D and Cowley, Anthony and Konkimalla, Ravi and Shivakumar, Shreyas S and Nguyen, Ty and Smith, Trey and Taylor, Camillo Jose and Kumar, Vijay},
  journal={IEEE Robotics and Automation Letters},
  volume={6},
  number={2},
  pages={2397--2404},
  year={2021},
  publisher={IEEE}
}

@inproceedings{downes2022city,
  title={City-wide street-to-satellite image geolocalization of a mobile ground agent},
  author={Downes, Lena M and Kim, Dong-Ki and Steiner, Ted J and How, Jonathan P},
  booktitle={2022 IEEE/RSJ International Conference on Intelligent Robots and Systems (IROS)},
  pages={11102--11108},
  year={2022},
  organization={IEEE}
}

@article{yu2025slim,
  title={Slim: Scalable and lightweight lidar mapping in urban environments},
  author={Yu, Zehuan and Qiao, Zhijian and Liu, Wenyi and Yin, Huan and Shen, Shaojie},
  journal={IEEE Transactions on Robotics},
  year={2025},
  publisher={IEEE}
}

@article{geiger2013vision,
  title={Vision meets robotics: The kitti dataset},
  author={Geiger, Andreas and Lenz, Philip and Stiller, Christoph and Urtasun, Raquel},
  journal={The international journal of robotics research},
  volume={32},
  number={11},
  pages={1231--1237},
  year={2013},
  publisher={Sage Publications Sage UK: London, England}
}

@inproceedings{shi2020beyond, title={Beyond Cross-view Image Retrieval: Highly Accurate Vehicle Localization Using Satellite Image}, author={Shi, Yujiao and Li, Hongdong}, booktitle={Proceedings of the IEEE Conference on Computer Vision and Pattern Recognition}, year={2022} }

@inproceedings{wang2024view,
  title={View from above: Orthogonal-view aware cross-view localization},
  author={Wang, Shan and Nguyen, Chuong and Liu, Jiawei and Zhang, Yanhao and Muthu, Sundaram and Maken, Fahira Afzal and Zhang, Kaihao and Li, Hongdong},
  booktitle={Proceedings of the IEEE/CVF Conference on Computer Vision and Pattern Recognition},
  pages={14843--14852},
  year={2024}
}

@inproceedings{kim2017satellite,
  title={Satellite image-based localization via learned embeddings},
  author={Kim, Dong-Ki and Walter, Matthew R},
  booktitle={2017 IEEE international conference on robotics and automation (ICRA)},
  pages={2073--2080},
  year={2017},
  organization={IEEE}
}

@article{lee2024autonomous,
  title={Autonomous vehicle localization without prior high-definition map},
  author={Lee, Sangmin and Ryu, Jee-Hwan},
  journal={IEEE Transactions on Robotics},
  volume={40},
  pages={2888--2906},
  year={2024},
  publisher={IEEE}
}

@article{chen2022direct,
  title={Direct lidar odometry: Fast localization with dense point clouds},
  author={Chen, Kenny and Lopez, Brett T and Agha-mohammadi, Ali-akbar and Mehta, Ankur},
  journal={IEEE Robotics and Automation Letters},
  volume={7},
  number={2},
  pages={2000--2007},
  year={2022},
  publisher={IEEE}
}

@article{hsieh2007adaptive,
  title={Adaptive teams of autonomous aerial and ground robots for situational awareness},
  author={Hsieh, M Ani and Cowley, Anthony and Keller, James F and Chaimowicz, Luiz and Grocholsky, Ben and Kumar, Vijay and Taylor, Camillo J and Endo, Yoichiro and Arkin, Ronald C and Jung, Boyoon and others},
  journal={Journal of field robotics},
  volume={24},
  number={11-12},
  pages={991--1014},
  year={2007},
  publisher={Wiley Online Library}
}

@inproceedings{zhang2024increasing,
  title={Increasing slam pose accuracy by ground-to-satellite image registration},
  author={Zhang, Yanhao and Shi, Yujiao and Wang, Shan and Vora, Ankit and Perincherry, Akhil and Chen, Yongbo and Li, Hongdong},
  booktitle={2024 IEEE International Conference on Robotics and Automation (ICRA)},
  pages={8522--8528},
  year={2024},
  organization={IEEE}
}

@inproceedings{wang2023satellite,
  title={Satellite image based cross-view localization for autonomous vehicle},
  author={Wang, Shan and Zhang, Yanhao and Vora, Ankit and Perincherry, Akhil and Li, Hengdong},
  booktitle={2023 IEEE International Conference on Robotics and Automation (ICRA)},
  pages={3592--3599},
  year={2023},
  organization={IEEE}
}

@article{frosi2023osm,
  title={Osm-slam: Aiding slam with openstreetmaps priors},
  author={Frosi, Matteo and Gobbi, Veronica and Matteucci, Matteo},
  journal={Frontiers in Robotics and AI},
  volume={10},
  pages={1064934},
  year={2023},
  publisher={Frontiers Media SA}
}

@article{przewodowski2024global,
  title={Global Localization using OpenStreetMap and Elevation Offsets.},
  author={Przewodowski, Andr{\'e} and Os{\'o}rio, Fernando Santos and Junior, Valdir Grassi},
  journal={J. Braz. Comput. Soc.},
  volume={30},
  number={1},
  pages={264--273},
  year={2024}
}

@article{li2026odometry,
  title={Odometry-Assisted LiDAR-OpenStreetMap Matching Method for Vehicle Global Positioning},
  author={Li, Zexing and Zuo, Runheng and Wang, Yafei and Ding, Fei and Wei, Chongfeng and Wu, Mingyu},
  journal={IEEE Internet of Things Journal},
  year={2026},
  publisher={IEEE}
}

@article{horn1987closed,
  title={Closed-form solution of absolute orientation using unit quaternions},
  author={Horn, Berthold KP},
  journal={Journal of the optical society of America A},
  volume={4},
  number={4},
  pages={629--642},
  year={1987},
  publisher={Optical Society of America}
}

@article{xia2025revisiting,
  title={Revisiting Cross-View Localization from Image Matching},
  author={Xia, Panwang and Wu, Qiong and Yu, Lei and Liu, Yi and Xiong, Mingtao and Liang, Lei and Zhang, Yongjun and Wan, Yi},
  journal={arXiv e-prints},
  pages={arXiv--2508},
  year={2025}
}

@article{munoz2024geo,
  title={Geo-localization based on dynamically weighted factor-graph},
  author={Mu{\~n}oz-Ba{\~n}{\'o}n, Miguel {\'A}ngel and Olivas, Alejandro and Velasco-S{\'a}nchez, Edison and Candelas, Francisco A and Torres, Fernando},
  journal={IEEE Robotics and Automation Letters},
  volume={9},
  number={6},
  pages={5599--5606},
  year={2024},
  publisher={IEEE}
}

@inproceedings{fervers2023uncertainty,
  title={Uncertainty-aware vision-based metric cross-view geolocalization},
  author={Fervers, Florian and Bullinger, Sebastian and Bodensteiner, Christoph and Arens, Michael and Stiefelhagen, Rainer},
  booktitle={Proceedings of the IEEE/CVF Conference on Computer Vision and Pattern Recognition},
  pages={21621--21631},
  year={2023}
}

@inproceedings{zhou2016fast,
  title={Fast global registration},
  author={Zhou, Qian-Yi and Park, Jaesik and Koltun, Vladlen},
  booktitle={European conference on computer vision},
  pages={766--782},
  year={2016},
  organization={Springer}
}

\end{document}